\documentclass{article}
\usepackage{spconf,amsmath,graphicx}
\pdfoutput=1

\title{MULTI-TASK LEARNING OF DEEP NEURAL NETWORKS FOR AUDIO VISUAL AUTOMATIC SPEECH RECOGNITION}
\name{Abhinav Thanda, Shankar M Venkatesan }
\address{Samsung R\&D Institute India, Bangalore}
\begin{document}
%
\maketitle
\begin{abstract}
Multi-task learning (MTL) involves the simultaneous training of two or more related tasks over
shared representations. In this work, we apply MTL to audio-visual automatic speech recognition(AV-ASR). 
Our primary task is to learn a mapping between audio-visual fused features and frame labels obtained from acoustic GMM/HMM model.
This is combined with an auxiliary task  which maps visual features to frame labels obtained from a separate visual GMM/HMM model.
The MTL model is tested at various levels of babble noise and the results are compared with a base-line hybrid DNN-HMM AV-ASR model. Our results indicate that MTL is especially useful at higher level of noise. Compared to base-line, upto 7\% relative improvement in WER is reported at -3 SNR dB\footnote{\label{version} September 12, 2016 version, submitted to ICASSP 2017}. 
\end{abstract}
\section{Introduction}
\label{sec:intro}
One way to build a noise robust Automatic Speech Recognition(ASR) system is to 
incorporate complementary information from a different modality that is independent
of noise in the speech. For example, in an audio-visual ASR(AV-ASR) system
visual information from the speaker's lip region when combined with audio inputs
have been shown to provide significant reduction in word error rates(WER)
when the audio modality is corrupted by noise. This was inspired by the fact that human perception of speech
is dependent on both auditory and visual senses as demonstrated by the famous McGurk effect\cite{mcgurk1976hearing}.

Traditionally, AV-ASR systems were implemented using GMM/HMM models \cite{dupont2000audio, bengio2004multimodal, huang2003improving, nefian2002dynamic}.
One way(called decision fusion method) is to model each modality by a separate GMM/HMM and at test time fuse the decisions of each stream by linearly combining the log-likelihoods to get the overall likelihood. In the feature fusion method, audio and visual features are combined usually by concatenation followed by a dimensionality reduction step. The fused features are modeled by a single GMM/HMM model.

With the advancement of deep learning based techniques in speech recognition\cite{dahl2012context}, corresponding AV-ASR
systems based on deep learning have been proposed \cite{huang2013audio, mroueh2015deep, ngiam2011multimodal, noda2015audio}. 
Deep learning based ASR systems tend to out-perform GMM/HMM models for several reasons. Firstly, GMM is a mixture model which acts as a "sum of experts" model, whereas DNN is a "product of experts". Also GMM/HMM systems require uncorrelated inputs and do not benefit from multiple frames of input whereas, this is not the case for a DNN\cite{hinton2012deep}.Correspondingly, deep learning based AV-ASR systems have been shown to perform better than their GMM/HMM counterparts. 

Similar to GMM/HMM feature fusion and decision fusion are possible for DNN-HMM systems\cite{potamianos2003recent, katsaggelos2015audiovisual, huang2013audio}. In decision fusion, each modality is modeled by a separate network. Mid-level fusion of features is also possible by fusing the hidden layer outputs of separate audio and visual features deep networks. 

So far, fusion models based on deep learning have been single task learning(STL) methods in which the fusion network predicts 
posterior probabilities of labels(usually context dependent tied HMM states) and the state space is common for both audio and 
visual modalities. An alternative approach, would be to assume different label set for the two modalities and train the network for two  tasks simultaneously using a shared representation. This is called multi-task learning(MTL)\cite{caruana1998multitask} and has been successfully applied to various problems of NLP\cite{collobert2008unified} and speech recognition\cite{chen2015multi,heigold2013multilingual,lin2009study,pironkovmulti} such as speech synthesis and multilingual acoustic modeling. One of the  tasks is considered to be a primary task while the other is an auxiliary task. The auxiliary task helps in better feature estimation in the hidden layers and proper generalization of the model. Better generalization results in improved robustness to noise. Once the MTL model is trained, the parameters corresponding to auxiliary task are usually discarded. In this work, we explore the application of MTL to AV-ASR systems. We model the visual stream by a separate GMM/HMM whose states are used as classes for the auxiliary task. We compare our method with a baseline DNN-HMM based model. We show that MTL gives better performance, especially at higher noise levels. 

The paper is organized as follows: Section 2 describes the feature extraction pipe-line for audio and visual features. MTL is explained
and is followed by a description of the primary and auxiliary tasks for AV-ASR. The training procedure and experimental results are  discussed in section 3. Section 4 discusses the relationship of our work with previous AV-ASR methods. Finally, we summarize
our work in section 5.

\section{MTL MODEL}
\label{sec:mtl}
\subsection{Feature Extraction}
\label{sec:featureextraction}

The sampling rate of audio data is converted to 16kHz. For each frame of speech
signal of 25ms duration, filter-bank features of 40 dimensions are extracted. Mean and variance normalization
is performed.

The video frame rate is increased to match the rate of audio frames through
interpolation. The region of interest(ROI) corresponding to the area surrounding speaker's mouth is extracted as follows:
Each frame is converted to gray scale and face detection is performed using Viola-Jones algorithm. The 64x64 lip region is extracted
by detecting 68 landmark points\cite{kazemi2014one} on the speakers face, and cropping the ROI surrounding speakers mouth and chin. 100 dimensional DCT features are extracted from the ROI. Similar to audio features, we perform mean and variance normalization.

\subsection{MTL-DNN}
\label{sec:mtl-dnn}
In MTL, a primary task along with one or more related secondary/auxiliary tasks are learnt simultaneously 
over shared representation of the data. The secondary task acts as a type of regularization and results 
in better generalization of the model. Once the MTL model is trained, the parameters corresponding to secondary 
task can be ignored. The cost function for an MTL network is given by

\begin{equation}\label{eq:1}
C_{mtl} = C_{main}+\sum_{n=1}^{N} \lambda_{n} C_{n} 
\end{equation}

where $ C_{mtl}, $ $C_{main}$, $C_{\lambda_{n}}$ represent cost functions  and $ 0 \le \lambda_{n} \le 1 $ for $n=1,2,...N$ are parameters indicating the relative importance of the secondary task with respect to the primary task. 

In this work, the primary as well as secondary tasks are hybrid DNN-HMM models. DNN-HMM hybrid systems are frame level classifiers and estimate the posterior probabilities of tied HMM states given an acoustic feature vector. The frame labels are obtained by
first training a bootstrap GMM/HMM model and then aligning the data against it.We train a tri-phone GMM/HMM acoustic model with Linear Discriminant Analysis(LDA) transformed audio features as input. The alignments obtained from the audio feature sequences and acoustic model are used as labels for the primary task of MTL-DNN Correspondingly, for secondary task we train another tri-phone GMM/HMM visual model with LDA transformed visual features. The visual input sequences are aligned against the visual GMM/HMM model. 

Here $N=1$ and \begin{equation}\label{eq:2}C_{main}= -\sum_{u=1}^{U}\sum_{t=1}^{T}log P_{a}(s_{ut}^{a}|O_{ut}^{av})\end{equation} and 
\begin{equation}\label{eq:3}C_{1}= -\sum_{u=1}^{U}\sum_{t=1}^{T}log P_{v}(s_{ut}^{v}|O^{av}_{ut})\end{equation} are the cross-entropy costs,
where $s_{ut}^{m}$ corresponds to output HMM state for an utterance $u$ at time $t$ for a given modality $m\in \{a,v\}$. $P_{a}$ and $P_{v}$ correspond to the soft-max outputs for each task.  $O_{ut}^{av}=[O_{ut}^{a},O_{ut}^{v}]$  corresponds to input fused features for an utterance $u$ at time $t$.

The MTL-DNN architecture used in this work, is shown in Figure 1. 40 dimensional audio features and 100 dimensional visual features are fused by simple concatenation. The 140 dimensional audio-visual features are then spliced with 10 frames which provide the contextual information.

The MTL-DNN is trained as follows: For the primary task, the inputs are audio-visual fused features in which both audio and visual modalities are present. In addition the primary task is also trained on fused features in which either audio or visual modality is suppressed by setting the corresponding features to very small values. For the secondary task, the inputs are fused features with audio modality suppressed i.e., the inputs correspond to only visual modality.

\begin{figure}[htb]
	
	\begin{minipage}[b]{1.0\linewidth}
		\centering
		\centerline{\includegraphics[width=6.5cm]{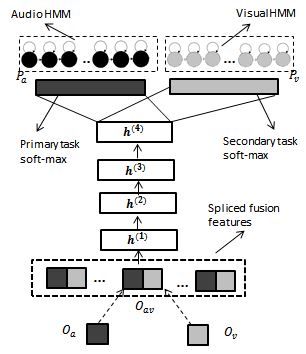}}
		\centerline{}\medskip
	\end{minipage}
	\caption{MTL-DNN architecture. The 40 dimensional audio(black) filterbank features are concatenated with 100 dimensional visual(gray) DCT features. The fused features
		are spliced with 10 context frames. Each hidden layer is of dimension 1500. The two soft-max layers correspond to audio and video tied HMM states }
	\label{fig:res}
\end{figure}


\section{EXPERIMENTS AND RESULTS}
\label{sec:results}

The system was trained and tested on GRID audio-visual corpus\cite{cooke2006audio}. The corpus is a collection of audio and video recordings of 34 speakers (18 male, 16 female) each uttering a 1000 sentences. Each utterance is approximately 3 seconds in length.  The syntactic structures of all sentences are similar\cite{cooke2006audio}. A simple language model following this structure was built for this work.

The dataset in effect consisted of 32692 utterances 90\% of the which (containing 29423 utterances) were used for training and cross validation while the remaining (10\%) data was used as test set. Both training and test data contain utterances from all of the speakers.  Models were trained and tested using Kaldi speech recognition tool kit\cite{povey2011kaldi} and Kaldi+PDNN\cite{miao2014kaldi+}. 

\textbf{Base-line} The base-line system is a single task learning feature fusion model(STL-DNN) somewhat similar to the network in \cite{huang2013audio}. Input to the network is 1540 dimensional vector. Feature extraction is same as described in \ref{sec:mtl-dnn}. The frame labels are obtained in the same way as the primary task of MTL-DNN described in \ref{sec:mtl-dnn}. The only difference between MTL-DNN and STL-DNN  is the absence of secondary task in STL-DNN.

 Pre-training is not performed as it did not provide any noticeable benefit. Cross-entropy loss function is minimized using mini-batch Stochastic Gradient Descent (SGD). The frames are shuffled randomly before each epoch. Batch size is set to 256 and initial learning rate is set to 0.008. New bob learning schedule is adopted, i.e., when the improvement in the cross-validation accuracy between two successive epochs falls below 0.5\%, the learning rate is halved.The halving is performed for each subsequent epoch until the increase in frame level accuracy is less than 0.1\%. At this point training is halted.

The models are tested with four levels of babble noise -3dB SNR, 0dB SNR, 10dB SNR and clean audio. Noise was added to test data artificially by mixing babble noise with clean audio files. At test time, the secondary task outputs are ignored.  Decoding is performed over a Weighted Finite State Transducer(WFST) built using GMM/HMM acoustic model, GRID corpus lexicon and the GRID language model. The model is tested once with audio-visual features and again with only audio features i.e., visual features set to small values. The posteriors are converted to log-likelihoods and passed to decoder which outputs the final word sequences. 

The results are tabulated in table 1. In line with previous AV-ASR systems, when visual modality is turned on, both MTL-DNN and STL-DNN provide significant gains in WER compared to audio only input. The large gains in our case are in part due to the small vocabulary and simple language model of GRID corpus. 

At low levels of noise (clean audio), the base-line system gives slightly better performance compared to both MTL-DNN models.However, as noise increases (0 dB and -3 dB)  the difference in performance improvement of MTL-DNN over STL-DNN becomes significant. At -3 SNR dB MTL-DNN with $\lambda=0.3$ gives nearly 7.23\% relative improvement (from 27.1 to 25.14) in WER over STL-DNN while $\lambda=0.1$ gives 1.6\% relative improvement. Among the two MTL-DNN models, the model with $ \lambda=0.3 $ gives better performance at 0 dB and -3 dB.  This is consistent with the idea that regularization due to auxiliary task reduces over-fitting which results in better performance with corrupted inputs. However, in our experiments we found that increasing $\lambda$ further, resulted in degradation of the performance of primary task.

The last row in table 1 corresponds to lip-reading, where audio modality is suppressed. Again MTL-DNN gives better performance compared to STL-DNN with approximately 3\% relative gain when $\lambda=0.3$ and 1.6\% relative gain when $\lambda=0.1$.
 	\begin{table}
 		\centering
 		\begin{tabular}{| c | c | c | c | c |}
 			\hline
 			\multicolumn{2}{ | c |}{Modality} & \multicolumn{3}{ c| }{WER \%} \\ \hline
 			Audio & Video & MTL $\lambda$=0.1 & MTL $\lambda$=0.3 & STL  \\ \hline
 			-3dB & OFF & 58.46 & 56.98 & 57.26 \\ \hline
 			-3dB & ON & 26.62 & 25.14 & 27.1 \\ \hline
 			0dB & OFF & 46.47 & 44.19 & 45.27 \\ \hline
 			0dB & ON & 17.87 & 16.59 & 18.03 \\ \hline
 			10dB & OFF & 10.65 & 9.93 & 10.72 \\ \hline
 			10dB & ON & 3.15 & 3.02 & 3.17 \\ \hline
 			CLEAN & OFF & 0.49 & 0.5 & 0.53 \\ \hline
 			CLEAN & ON & 0.52 & 0.45 & 0.43 \\ \hline
 			OFF & ON & 9.10 & 8.98 & 9.25 \\ \hline
 			
 		\end{tabular}
 		\label{table:WERCOMPARE}
 		\caption[short text]{\% WER comparison between MTL-DNN and STL-DNN. In case of MTL-DNN two values of $\lambda$ are reported.}
 	\end{table}

\section{RELATION TO PRIOR WORK}
\label{sec:prior}
 
Multi-task learning(MTL) has been successfully applied to various problems\cite{pironkovmulti, chen2015multi} in NLP\cite{collobert2008unified}, speech synthesis\cite{wu2015deep}, multilingual acoustic modeling\cite{lin2009study, heigold2013multilingual} and other ASR applications \cite{chen2015multi, pironkovmulti}. 
The application of MTL to AV-ASR was suggested in \cite{chen2015multi} although no results were reported. Our training procedure to some extent similar to \cite{heigold2013multilingual}. Like \cite{heigold2013multilingual} we separate the inputs and outputs of two tasks. 

Our work is inspired by \cite{ngiam2011multimodal}  and \cite{huang2013audio}. In \cite{ngiam2011multimodal} the authors employ unsupervised learning methods to obtain a shared representations of the audio and visual modalities which are then used in a separate supervised training step. In our work, learning the shared representation as well as supervised training are taken care by multi-task learning. In contrast to \cite{huang2013audio} we employ low-level feature fusion. During training we suppress one of the modalities to ensure that the network does not over-fit to a particular modality. In addition, our results are reported on GRID corpus which includes digits, alphabets and words different from the continuous digit dataset used in \cite{huang2013audio}.

\section{CONCLUSIONS AND FUTURE WORK}
\label{sec:refs}

In this work, we applied multi-task learning to audio-visual speech recognition. We described the primary and auxiliary tasks to train MTL model. We proposed a network architecture and training protocol for the network. The model was compared with baseline STL-DNN model at various levels of babble noise. Our results indicate that MTL results in the improvement of WER over baseline model, especially at higher levels of noise. In our future work, we would like to compare MTL with ensemble learning for AV-ASR.

\bibliographystyle{IEEEbib}
\bibliography{strings,refs}

\end{document}